\documentclass{article}

\PassOptionsToPackage{numbers, compress}{natbib}

\usepackage[preprint]{neurips_2019}

\usepackage[utf8]{inputenc} 
\usepackage[T1]{fontenc}    
\usepackage{hyperref}       
\usepackage{url}            
\usepackage{booktabs}       
\usepackage{amsfonts}       
\usepackage{nicefrac}       
\usepackage{microtype}      

\usepackage{color}
\usepackage[normalem]{ulem}
\usepackage{amsmath}
\usepackage{amssymb}
\usepackage[ruled]{algorithm2e}
\usepackage[pdftex]{graphicx}
\usepackage{subcaption}
\usepackage{wrapfig}

\newcommand\blfootnote[1]{%
  \begingroup
  \renewcommand\thefootnote{}\footnote{#1}%
  \addtocounter{footnote}{-1}%
  \endgroup
}

\newcommand{\rvauthcommands}[4]{%
\expandafter\newcommand\csname #1rev\endcsname{#4} 
\expandafter\newcommand\csname #1\endcsname[2][0]{\ifnum##1>\csname #1rev\endcsname\textcolor{#3}{##2}\else##2\fi} 
\expandafter\newcommand\csname #1t\endcsname[2][0]{\ifnum##1>\csname #1rev\endcsname\textcolor{#3}{\sout{##2}}\fi} 
\expandafter\newcommand\csname #1r\endcsname[3][0]{\ifnum##1>\csname #1rev\endcsname\textcolor{#3}{\sout{##2}##3}\else##3\fi} 
\expandafter\newcommand\csname #1c\endcsname[2][0]{\ifnum##1>\csname #1rev\endcsname\textcolor{#3}{\scriptsize{/\uline{##2}/}}\fi} 
}

\rvauthcommands{yh}{yh}{blue}{-1}
\rvauthcommands{ml}{mlee}{red}{-1}
\rvauthcommands{jy}{jinyoung}{magenta}{-1}

\title{Neuro-Optimization: Learning Objective Functions Using Neural Networks}

\author{
  Younghan Jeon$^{\dagger*}$ \hspace{5em} Minsik Lee$^{\ddagger*}$ \hspace{5em} Jin Young Choi$^{\dagger}$ \\
  \texttt{yh1992@snu.ac.kr} \hspace{1em} \texttt{mleepaper@hanyang.ac.kr} \hspace{1em} \texttt{jychoi@snu.ac.kr}\\
  Department of Electrical and Computer Engineering, ASRI, Seoul National University$^{\dagger}$ \\
  Division of Electrical Engineering, Hanyang University$^{\ddagger}$ \\
}
\begin{document}

\maketitle

\begin{abstract}
Mathematical optimization is widely used in various research fields.
With a carefully-designed objective function, mathematical optimization can be quite helpful in solving many problems.
However, objective functions are usually handcrafted and designing a good one can be quite challenging.
In this paper, we propose a novel framework to learn the objective function based on a neural network.
The basic idea is to consider the neural network as an objective function, and the input as an optimization variable.
For the learning of objective function from the training data, two processes are conducted: In the inner process, the optimization variable (the input of the network) are optimized to minimize the objective function (the network output), while fixing the network weights.
In the outer process, on the other hand, the weights are optimized based on how close the final solution of the inner process is to the desired solution.
After learning the objective function, the solution for the test set is obtained in the same manner of the inner process.
The potential and applicability of our approach are demonstrated by the experiments on toy examples and a computer vision task, optical flow.

\blfootnote{$^*$ Authors contributed equally.}
\end{abstract}
\section{Introduction}
\label{intro}

Mathematical optimization has been vastly used in many research areas for decades. It has been one of the core parts in solving many important problems and has given us many insights about the problems themselves. In utilizing mathematical optimization, designing a good objective function, sometimes denoted also as a cost or loss function, is often one of the most important parts. A particular choice of the objective function can affect the convergence behavior as well as the final solution, and naturally, the performance in an actual application.

For example, there have been suggested a lot of objective functions to find optical flow, i.e., the motion of pixels between two consecutive image frames, in computer vision area~\cite{horn, flow2, flow4, flow5}.
Many of them are based on the widely-used Horn-Schunck (HS) energy function ~\cite{horn}.
The usual implementation of the HS method is based on an approximated version of the original objective function, i.e.,
\begin{equation}
\label{eq1}
    E(I_1, I_2, u, v) = \iint \left[ \left(I_1(x,~y)-I_2(x+u,~y+v) \right)^{2} +\lambda \left(\left\Vert\nabla u \right\Vert^{2}+\left\Vert\nabla v \right\Vert^{2} \right) \right]~\textrm{d}x\textrm{d}y,
\end{equation}
where $x$ and $y$ are horizontal and vertical coordinates, respectively,
$I_1$ and $I_2$ are the two images, and $(u, v)$ is an optical flow vector at pixel $(x, y)$.
The first term $(I_1(x,~y)-I_2(x+u,~y+v))^{2}$ represents the photometric consistency and the second term implies the similarity of flow vectors between nearby pixels.
This manually designed energy function has been modified by numerous studies to overcome its poor performance.
However, many of these modifications were designed heuristically and it is difficult to know which is exactly better.

The motivation of this paper is what if we learn the objective function itself instead of manually designing it.
In most cases, objective functions are handcrafted with some unrealistic assumptions, heuristics, and approximations, and they are constantly modified by trial-and-error.
Thus, it is hard to validate the effectiveness of a given design and is also difficult to compare different choices, because in most cases it requires a fair amount of actual evaluations.
Moreover, there are usually a number of hyper-parameters such as $\lambda$ in equation (\ref{eq1}) that must be fine-tuned, so it is difficult to guarantee consistent performance in various environments.
In this paper, we propose a novel framework to learn objective functions to avoid such difficulties.
The purpose of our method is to learn the objective function by a neural network so that the optimization variable, which becomes the input of the network, gets closer to the desired value after optimizing the objective function.

Contributions of the paper are summarized as follows.

\begin{itemize}
\item We propose a novel framework so-called Neuro-Optimization which learns the objective function of a particular problem by a neural network and obtains the solution by optimizing it. This method can be applied to various problems that the desired solution is given for each training data.
\item Adopting our method for optical flow is introduced as an example application. This gives an illustration of how to apply our method to real problems.
\item Experiments show that our method can learn good objective functions for a variety of practical tasks based on an appropriate neural network.
\end{itemize}
\section{Related Works}
\label{works}

\textbf{Optimization:} 
Mathematical optimization based on objective functions are importantly used in many fields.
There have been lots of studies on optimization techniques to find an optimal value for a given objective function~\cite{adam, adagrad, adadelta, rmsprop}.
Recently, several researches have emerged to find an efficient optimization strategy via machine learning~\cite{learntolearn, learnoptnn}.
These works are loosely related with the proposed method, but their goal, finding a good optimizer for a given objective function, is different from ours.

\textbf{Inverse reinforcement learning:} 
Studies on inverse reinforcement learning are also somewhat related to our method ~\cite{irl1, irl2, irl3, irl4}.
The goal of reinforcement learning is to learn the policy that maximizes the sum of all rewards.
On the other hand, inverse reinforcement learning aims to find a good reward function that explains a given policy well.
Inverse reinforcement learning and the proposed method has many similarities in that both solves some sort of inverse problems, but they differ in terms of subjects of interest.

\textbf{Optical flow:}
As briefly described in the introduction, optical flow is one of the major branches of computer vision which aims to acquire motions by matching pixels in two images.
Significant progress has been made since the study of Horn and Schunck~\cite{horn}, and many recent studies are based on this.
They proposed an objective function consisting of a data term and a smoothness (regularization) term to solve optical flow.
Since the performance was not so good at the time, various studies were conducted to improve either the objective function or the optimization method~\cite{flow1, flow2, flow3, flow4, secret}.
Although performance has improved considerably thanks to these studies, their modifications solve only some partial issues by heuristics and a number of hyper-parameters must be fine-tuned.
Moreover, it is hard to compare the superiority between the objective functions.
Meanwhile, it has not been long since the introduction of neural networks in the optical flow area.
Beginning with FlowNet~\cite{flownet} and FlowNet2~\cite{flownet2}, many researches have been carried out to apply neural network to optical flow~\cite{pwcnet, litenet}.
These methods have relatively less computational loads and they have recently improved the level of performance greatly, but designing an overall structure is very difficult and requires many empirical experiences.
Unlike the traditional methods, they obtain optical flow directly from the output of the network without any energy function.
Compared to the existing methods, our method can be considered as a blend between the traditional approach and the neural-network-based approach.
In the proposed framework, optical flow can be estimated by optimizing the learned energy function which may contain good properties that are not yet discovered.
\section{Proposed Method}
\label{method}

In this section, we describe the problem setting and the basic idea of the proposed method. The proposed method learns an objective function from a set of training data. Based on the desired solution of each sample case, the weights of the neural network, i.e., the objective function, is updated. Based on the fact that the proposed method infers the objective function from the desired solutions, it can be considered as a type of inverse process of mathematical optimization. In this regard, the proposed method has some similarities to the inverse reinforcement learning~\cite{irl2}.

\subsection{Problem Setting}

Suppose that we want to learn an objective function from a set of problems of the same kind. For example, we might be interested in solving a 3D human pose estimation problem where the corresponding 2D pose is given, and attempt to formulate it as an optimization problem. Then, each 2D pose example can define a different problem, where the basic form of the objective function is shared across different examples. Unfortunately, we are not yet to know the exact form, but instead we have the desired solution for each problem instance. To formally represent the situation, let us assume that there are a set of training samples $S = \{(\mathbf{a}_1, \mathbf{t}_1), (\mathbf{a}_2, \mathbf{t}_2), \cdots, (\mathbf{a}_M, \mathbf{t}_M)\}$ where $\mathbf{a}_i$ represents the distinctive particular information about the $i$-th problem instance, $\mathbf{t}_i$ is the corresponding desired solution, and $M$ represents the number of training instances. In the above scenario, each 2D pose and its ground truth 3D pose become $\mathbf{a}_i$ and $\mathbf{t}_i$, respectively. Based on this training set, our goal is to learn the form of the objective function.

In the proposed framework, there are two optimization processes; the inner process and the outer process. The inner process is an optimization process where a neural network (a scalar output) itself becomes the objective function and the variable to adjust is the input of the network. In this process, the weights of the network are fixed. On the other hand, the outer process is the actual process that learns the objective function. There are several inner processes inside an outer process, and each inner process is conducted on a different training sample. Based on how close the final solution of each inner process is to the desired solution, the loss of the outer process is defined. Based on this loss, the weights of the network, i.e., the objective function, is updated in the outer process. To distinguish the terms, we will denote the objective function of the inner process as the \emph{inner objective} and the loss of the outer process as the \emph{outer loss}. Then, the inner objective is the function that we want to learn, and the outer loss is the error from the desired solution to learn the inner objective.

In the following sections, the inner objective is denoted by a neural network $f(\mathbf{x}, \mathbf{a}; \theta)$. Here, $\mathbf{x}$ is one of the input and is the variable to adjust in the inner process. This is called the \emph{inner variables} or the \emph{variable input} hereafter. $\mathbf{a}$, another input of $f$, is the particular information for a specific training example, as described above, and is fixed throughout an inner process. Accordingly, this is called the \emph{coefficients} or the \emph{sample input} in this paper. Lastly, $\theta$ represents the weights of the neural network. $\theta$ is adjusted during the outer process, so it is named the \emph{outer variables} or the \emph{parameters}. Note that in the above definitions, all the first names represent the roles for the two processes of the proposed framework, while the second names describe those in the neural network. The goal of the proposed framework is to adjust $\theta$ so that $\mathbf{x}$ reaches $\mathbf{t}$, the \emph{desired value} or \emph{target}, in each inner process. Note that $\mathbf{x}$, $\mathbf{a}$, and $\mathbf{t}$ are generally assumed to be vectors in this paper, however, other types can be similarly explained.

\begin{figure}[t]
  \centering
  \includegraphics[width=0.9\linewidth]{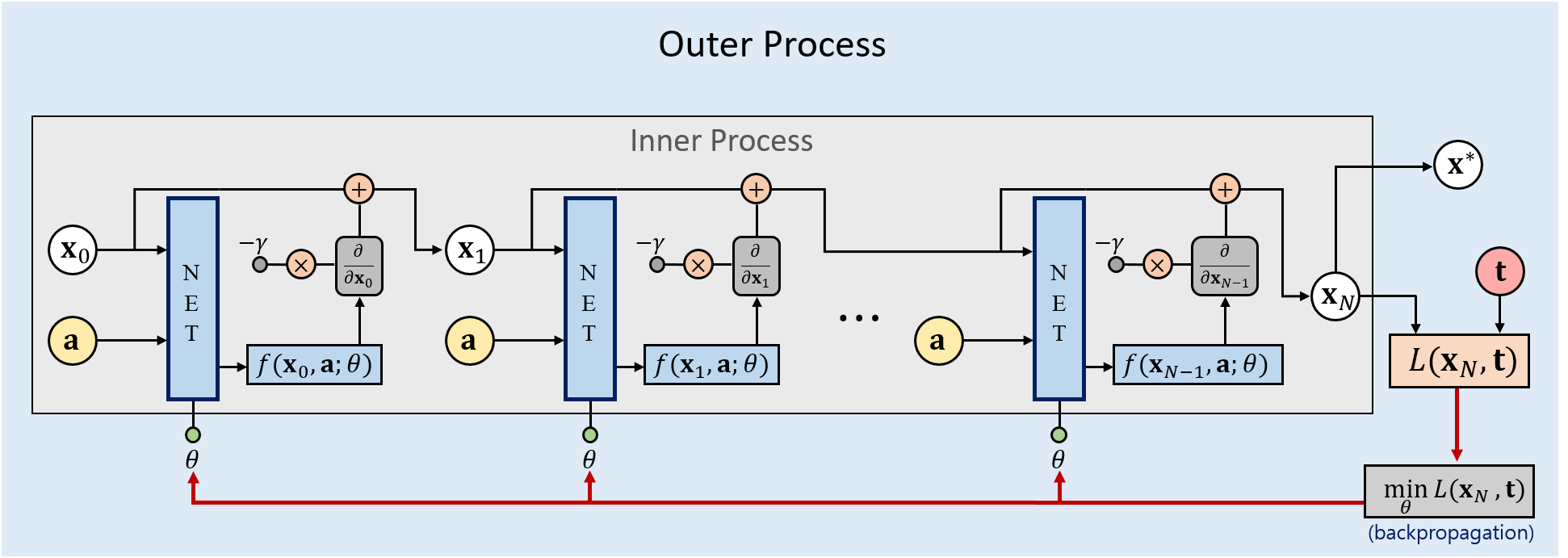}
  \caption{ {\bf{Neuro-Optimization:}} the overall scheme with mini-batch size $h=1$. For training, $\mathbf{x}_0$ is optimized to $\mathbf{x}_N$ by gradient descent for a given sample input $\mathbf{a}$ in the inner process, and the network weights $\theta$ are optimized to reduce the dissimilarity $L(\mathbf{x}_N, \mathbf{t})$ of $\mathbf{x}_N$ from the desired solution $\mathbf{t}$ in the outer process. Note that all $\mathbf{x}_n$ are produced based on the gradient of $f$, so $\mathbf{x}_N$ is a function of $\theta$ (as well as $\mathbf{x}_0$ and $\mathbf{a}$). In the test phase, only an inner process is needed to obtain a solution $\mathbf{x}^*$.}
  \label{fig1}
  \vspace{-0.3cm}
\end{figure}

\subsection{Algorithm}

As explained earlier, the proposed method consists of the inner and outer processes. The inner process computes the changes of the variable input $\mathbf{x}$, given a particular sample input $\mathbf{a}$. The process itself is an optimization process, and we assume that a simple gradient descent is used. Let us assume that $\mathbf{x}_n$ represents the variable input after the $n$-th iteration in the inner process, then the next variable input is calculated as
\begin{equation}
\label{eq2}
    \mathbf{x}_{n+1} = \mathbf{x}_n - \gamma\nabla_{\mathbf{x}_n} f(\mathbf{x}_n, \mathbf{a}; \theta),
\end{equation}
starting from an initial value $\mathbf{x}_0$. Here, $\gamma$ represents the step size. In this paper, $\gamma$ is fixed and considered as a hyper-parameter. We might consider a fancy method, such as line search algorithms or the Barzilai-Borwein method~\cite{BB}, to determine $\gamma$ in each iteration, but it is left as future work to focus on the main idea. Similarly, we might also consider a more complicated update rule such as Adam~\cite{adam} instead of the basic gradient descent, however, it is beyond the scope of this paper. The inner process repeats the above update rule $N$ times, of which the number is also fixed in the proposed method.

The outer process utilizes the collected results of several inner processes performed for different training samples $\mathbf{a}$.
Then, for the result of each inner process, the corresponding loss $L(\mathbf{x}_{N}, \mathbf{t})$ is defined between $\mathbf{x}_{N}$ and $\mathbf{t}$ for some appropriate distance or metric $L$, e.g., mean squared error. The average of these losses becomes the outer loss, and based on this, $\theta$ can be updated using existing optimizers for neural networks. Note here that $\mathbf{x}_N$ is the result of gradient descent based on $f(\mathbf{x}, \mathbf{a}; \theta)$, so it is actually a function of $\theta$ as well as $\mathbf{x}_0$ and $\mathbf{a}$. The entire procedure is summarized in Algorithm \ref{alg2}.

\begin{algorithm}
\label{alg2}
\caption{Neuro-Optimization}
{\bf Input:} Training set $S$, mini-batch size $h$, step size $\gamma$, number of steps $N$\\
{\bf Initialize:} Network parameters $\theta$\\
\For{number of training epochs}
{
    Initialize $\bar{B} = \{ B | B \text{ is a partition of } S \text{ with size } h \}$. \\
    \For{$B \in \bar{B}$}
    {
        Initialize $Y$ to an empty set. \\
        \For{$(\mathbf{a}, \mathbf{t}) \in B$}
        {
            Initialize $\mathbf{x}_0$.\\
            \For{$n=0,1, \dots, N-1$}
            {
                $\mathbf{x}_{n+1} = \mathbf{x}_n - \gamma \cfrac{\partial f(\mathbf{x}_n, \mathbf{a}; \theta)}{\partial \mathbf{x}_n}$.\\
            }
            Add $(\mathbf{x}_N, \mathbf{t})$ to $Y$.
        }
        Update $\theta$ to minimize  $\frac{1}{h} \sum_{(\mathbf{x}_N, \mathbf{t}) \in Y} L(\mathbf{x}_N, \mathbf{t})$.
    }
}
\end{algorithm}

Here, the initialization of $\mathbf{x}_0$ can be important for the success of the proposed method, as in many problems of mathematical optimization. The appropriate choice depends on the particular problem, which will be explained later. After training is finished, test procedure can be easily conducted. Given a new sample input $\mathbf{a}'$, we can simply perform an inner process to find the solution. Figure~\ref{fig1} shows the overall scheme of the proposed algorithm.

There are a few things worth noting about the proposed method. First of all, our method might seem similar to some existing networks approximating the unrolled steps of an optimization procedure~\cite{unroll1, unroll2, learntolearn}. If we consider the path from $\mathbf{x}_0$ to $\mathbf{x}_N$ including the gradient operations, i.e., the very procedure of an inner process, as a neural network, then there is indeed a similarity. However, in these networks, the direction is somewhat reversed: A few steps of update equations in an optimization procedure are derived and they are approximated to similar neural network layers with variable weights. Due to this approximation, the equation representing each layer is not a valid update form any more because some of the terms have been relaxed to variables. In other words, these networks are more of newly designed network structures inspired by optimization procedures rather than actual ones. As a result, they do not correspond to a specific objective function. In contrast to these networks, our inner process remains as an optimization procedure and the proposed method learns a valid objective function. In fact, as depicted in Figure~\ref{fig1}, our method can also be viewed as a type of recurrent neural network (RNN), since the inner process repeats the same iteration step several times, and an iteration step involves the result of the previous iteration.

Another thing to consider is that the proposed method needs the ability to calculate the second derivative, i.e., there are gradient operations with respect to (w.r.t.) $\mathbf{x}$ in the inner process and the outer process also need gradients w.r.t. $\theta$. A fortunate fact is that we do not need the entire Hessian in all cases. The reason is two-fold: The only second derivative we need is $\frac{\partial^2 f}{\partial \mathbf{x} \partial \theta}$ and not $\frac{\partial^2 f}{\partial \mathbf{x}^2}$ or $\frac{\partial^2 f}{\partial \theta^2}$. Moreover, even though the inner process involves some gradients, they are summed up to a scalar in calculating $L(\mathbf{x}_N, \mathbf{t})$. This means that, in most cases, we do not exactly need the ability to calculate the Hessian but that to calculate the product of the Hessian and some vector. In practice, the proposed method has operated with no problem in popular deep learning libraries, such as TensorFlow and PyTorch.

Compared to a handcrafted objective function, the objective function learned by the proposed method can be better in regard to performance. If $L$ is chosen to be the same metric for evaluating performance, then there is a chance to learn an objective function that maximizes or minimizes the given metric. This is a nice characteristic of the proposed method since handcrafted objective functions do not usually align exactly with the corresponding performance metrics.

\subsection{Application}
\label{3.3}

The proposed framework has a lot of potential in that it can be applied to various applications. It is well-known that it is possible for neural networks to approximate variety of functions, according to the universal approximation theorem~\cite{uat1, uat2}, if we have enough resources such as computation power, memory, and training data. However, in reality, resources are limited and it requires some efforts to apply the proposed method efficiently in practical applications. Most importantly, one has to consider the characteristics of the application and choose an adequate structure to reduce the search space. For example, one can design a network structure that is similar to, but more general than, an existing handcrafted objective function for a particular application. In this section, we will discuss an example application to optical flow for demonstration.

To apply the proposed method to optical flow, we design a neural network for the inner objective utilizing the basic form of the HS energy function (\ref{eq1}), instead of using a black-box neural network. Let us assume that $J \triangleq g(I; \theta)$ is a feature extraction network for images, where its result $J$ is an image with the same width and height as $I$, but with different channels. Then the inner objective for optical flow is defined as follows:
\begin{equation}
\label{eq2}
    \begin{split}
    f_\text{op}(&u, v, I_1, I_2; \theta) \triangleq E(u, v, J_1, J_2) = \\ &\iint \left[ \left(J_1(x,~y)-J_2(x+u,~y+v) \right)^{2} +\lambda \left(\left\Vert\nabla u \right\Vert^{2}+\left\Vert\nabla v \right\Vert^{2} \right)\right]~\textrm{d}x\textrm{d}y,
    \end{split}
\end{equation}
where $J_1 \triangleq g(I_1; \theta)$ and $J_2 \triangleq g(I_2; \theta)$. Hence, this is basically an HS energy function applied to feature images extracted by a neural network. An important thing here is that the feature extraction network itself will be trained during the procedure of the proposed method. In this paper, U-net~\cite{unet} is used as $g$, with additional padding operations to maintain the size of an image. U-net is based on a pyramid-like structure, which can be helpful for finding optical flow as long discussed in the literature. In the proposed framework, $(u, v)$, the optical flow vector, are the inner variables, and $I_1$ and $I_2$, the images, are the coefficients. Note that these terms are shown in the above equation as if they are 2-dimensional functions, but in reality, they are handled as 2-dimensional arrays with the same size in the proposed method. Accordingly, $J_2(x+u,~y+v)$ in the above equation is regarded as a warping operation, where the bilinear interpolation is used. Likewise, the image gradients in the above equation are approximated based on finite differences.

For this problem, the training set $K$ contains $M$ bundles of $(I_1, I_2, u_{gt}, v_{gt})$, where $u_{gt}$ and $v_{gt}$ are ground truth optical flow for $I_1$ and $I_2$. The average end point error (EPE)~\cite{sintel} is used as the outer loss, i.e., $L_\text{op}(u, v, u', v') = \frac{1}{V} \iint \sqrt{(u-u')^2 + (v-v')^2}~\textrm{d}x\textrm{d}y$ where $V$ is the area of the images (of course, $L_\text{op}(u, v, u', v')$ is also computed based on discrete approximation). Algorithm \ref{alg3} summarizes the whole procedure of the proposed method for optical flow.

\begin{algorithm}
\label{alg3}
\caption{Neuro-Optimization for Optical Flow}
{\bf Input:} Training set $K$, mini-batch size $h$, step size $\gamma$, number of steps $N$, regularization constant $\lambda$\\
{\bf Initialize:} Network parameters $\theta$\\
\For{number of training epochs}
{
    Initialize $\bar{B} = \{ B | B \text{ is a partition of } S \text{ with size } h \}$. \\
    \For{$B \in \bar{B}$}
    {
        Initialize $Y$ to an empty set.\\
        \For{$(I_1, I_2, u_{gt}, v_{gt}) \in B$}
        {
            Initialize $u_0$ and $v_0$.\\
            \For{$n=0,1, \dots, N-1$}
            {
                $J_1 = g(I_1, \theta)$.\\
                $J_2 = g(I_2, \theta)$.\\
                $u_{n+1} = u_n - \gamma \cfrac{\partial E(u_n, v_n, J_1, J_2)}{\partial u_n}~$.\\
                $v_{n+1} = v_n - \gamma \cfrac{\partial E(u_n, v_n, J_1, J_2)}{\partial v_n}~$.\\
            }
            Add $(u_N, v_N, u_{gt}, v_{gt})$ to $Y$.
        }
        Update $\theta$ to minimize $\frac{1}{h} \sum_{(u_N, v_N, u_{gt}, v_{gt}) \in Y} L_\text{op}(u_{N}, v_{N}, u_{gt}, v_{gt})$
        
    }
}
\end{algorithm}

Since optical flow is a complicated problem, having good initial points $u_0$ and $v_0$ can be vital for the success of the proposed method. If we do not have good initial points, then the possible ranges of $u$ and $v$ that the proposed method has to learn the objective function for can be too large. To avoid this issue, we use existing algorithm, PWC-net~\cite{pwcnet}, as an initializer. In this regard, the product of the proposed method can be viewed as a post-processing scheme to improve the performance further.

\subsection{Learning Hyper-parameters}
\label{3.4}
When designing the objective function, other traditional methods spend a lot of time tuning hyper-parameters.
On the other hand, our algorithms are less sensitive since the network automatically learns a proper objective function.
Moreover, we can also attempt to learn the hyper-parameters in the proposed framework, considering them as network parameters.
For example, the regularization constant $\lambda$ in (\ref{eq2}) can be included in $\theta$ as a parameter.
Although omitted in Section \ref{sec:exp}, we have actually tried training $\lambda$ several times in this way and have fixed $\lambda$ to the converged value for every other training attempts.
Hyper-parameters can be easily chosen in this way or it is also possible to learn them together with other network parameters in each iteration.
\section{Experiments}
\label{sec:exp}
The proposed method has been experimented on some toy examples and optical flow estimation.
All experiments have been conducted based on PyTorch with an Nvidia Titan X GPU.
We did not use any additional techniques beyond the scope of this paper to improve performance.

\subsection{Toy Examples}
In this section, the feasibility of our algorithm is demonstrated for the following four toy examples. Note that the functional relations of the problems are not known, but only the training samples $(\mathbf{a}, \mathbf{t})$ are given. Then, the proposed method learns the form of the objective function in the training phase so that optimizing the objective function for each sample input $\mathbf{a}$ yields $\mathbf{x}^* \approx \mathbf{t}$. Accordingly, in the test phase, we optimize the trained objective function for a new sample input and compare the result to the ground truth target.


\begin{figure}[t]
  \centering
    \begin{subfigure}[t]{.45\linewidth}
		\centering
		\includegraphics[width=1\linewidth]{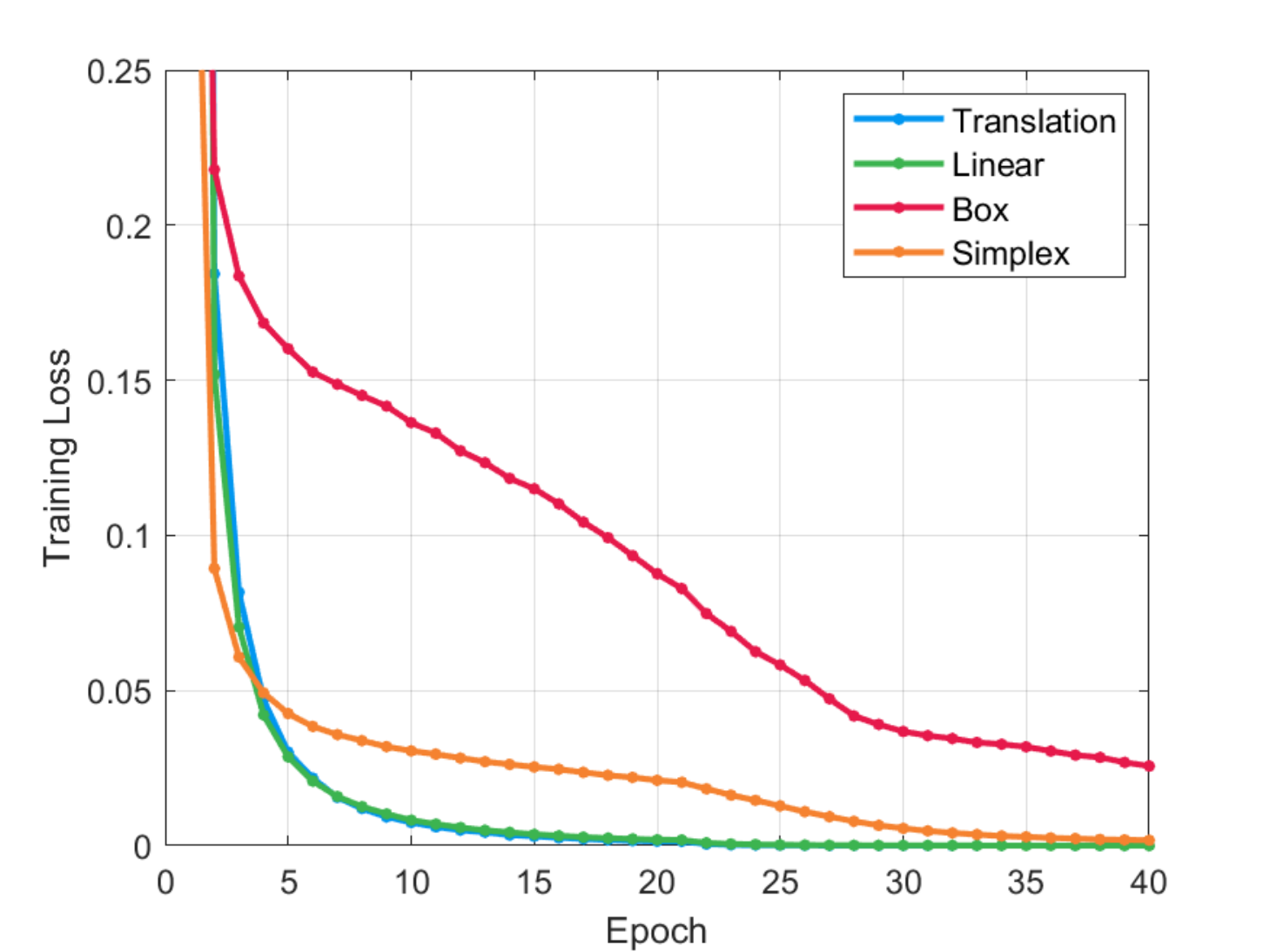}
		\caption{Training loss per epoch}
		\label{sfig:testa}
	\end{subfigure}
	\begin{subfigure}[t]{.45\linewidth}
		\centering
		\includegraphics[width=1\linewidth]{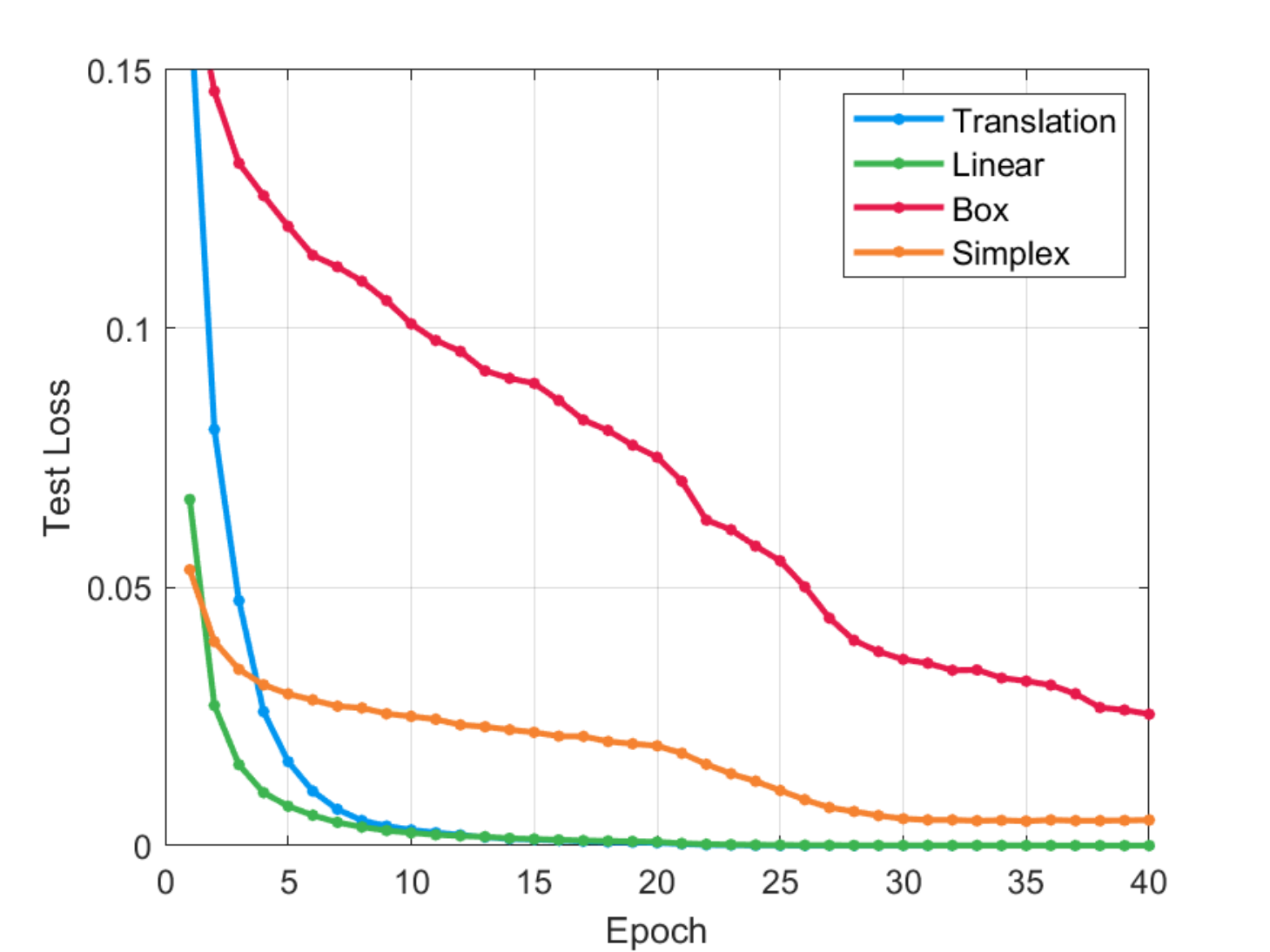}
		\caption{Test loss per epoch}
		\label{sfig:testa}
	\end{subfigure}
	\begin{subfigure}[t]{.45\linewidth}
		\centering
		\includegraphics[width=1\linewidth]{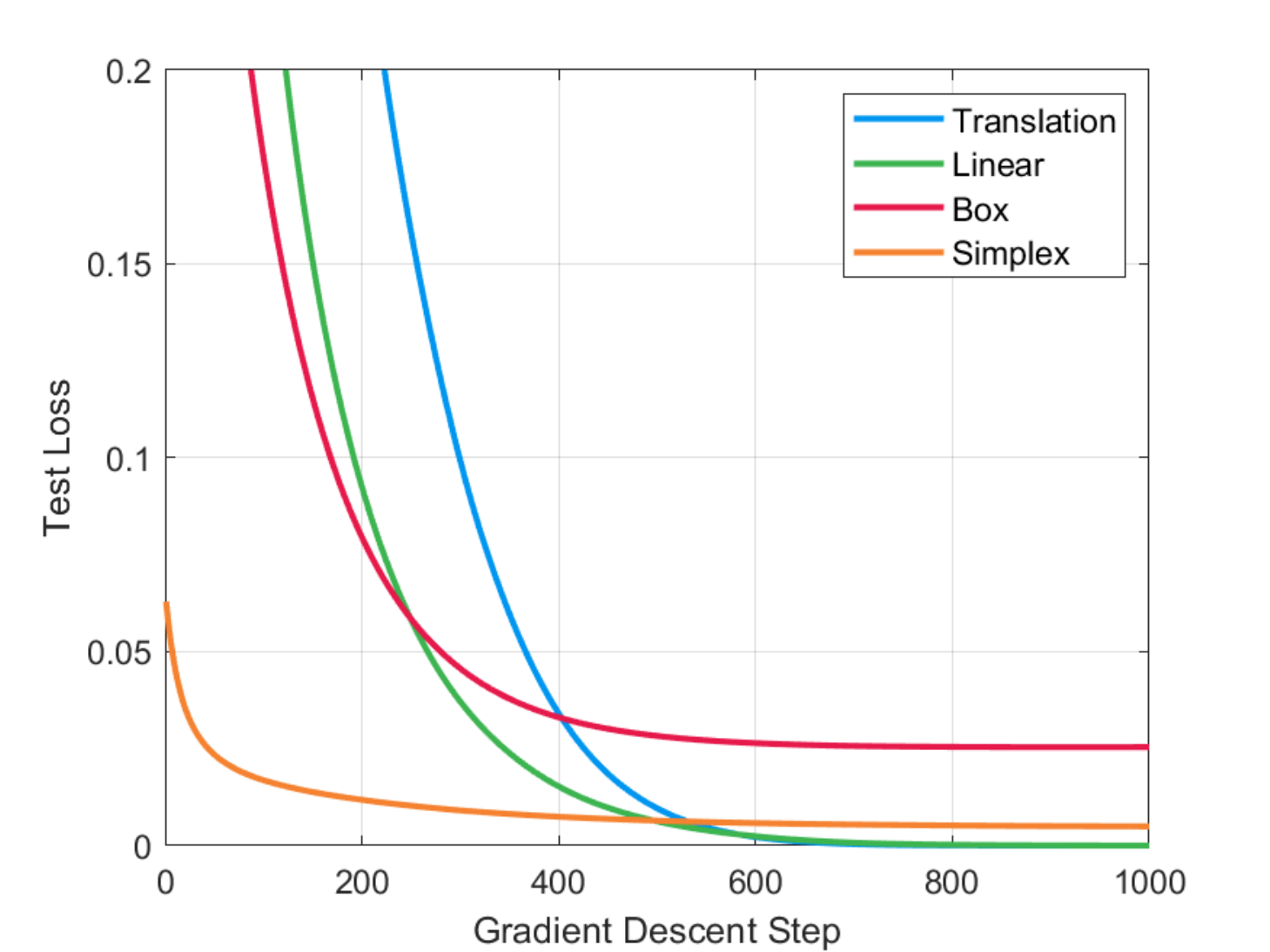}
		\caption{Test loss per gradient descent step}
		\label{sfig:testa}
	\end{subfigure}
	\begin{subfigure}[t]{.45\linewidth}
		\centering
		\includegraphics[width=1\linewidth]{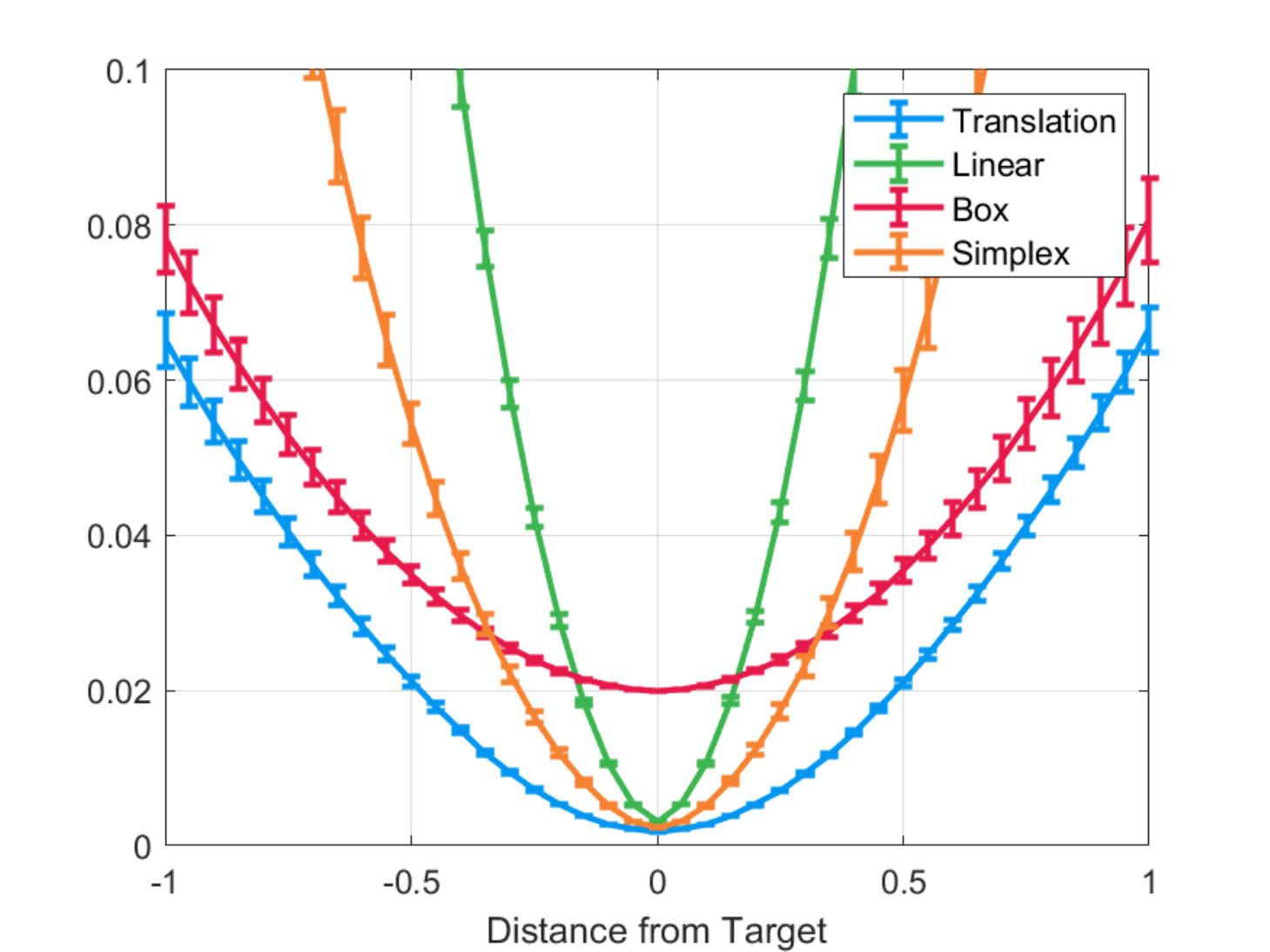}
		\caption{Objective function}
		\label{sfig:testa}
	\end{subfigure}
  \caption{\textbf{Experimental Results for Toy Examples.}
  (a) and (b) show the training loss and the test loss, respectively, for training epochs. (c) shows that the test loss decreases along the gradient descent steps (inner process). (d) shows the shape of learned objective function near the target.}
  \label{fig2}
  \vspace{-0.3cm}
\end{figure}

{\bf Translation:}
This is the simplest toy example where the desired solution for a sample input $\mathbf{a}$ is $\mathbf{t} = \mathbf{a} + \mathbf{d}$ for some vector $\mathbf{d}$.
$\mathbf{d}$ is randomly picked and fixed for the whole process. One of ideal objective functions for this problem is $\| \mathbf{x} - (\mathbf{a} + \mathbf{d}) \|^2$.

{\bf Linear equation:}
Here we consider the case of linear equation $W\mathbf{x}=\mathbf{a}$ where $W$ is a $10 \times 10$ matrix with full rank.
It can be considered as a convex optimization problem $\min_{\mathbf{x}}\left\Vert W\mathbf{x}-\mathbf{a}\right\Vert^2$.
The desired value should be $\mathbf{t} = W^{-1}\mathbf{a}$.

{\bf Constrained problem 1. Box constraint:}
This is a box-constrained problem $\min_{-\mathbf{1}\le\mathbf{x}\le \mathbf{1}}\left\Vert\mathbf{x}-\mathbf{a}\right\Vert^2$. Here, the inequalities are element-wise. This can be reformulated as an unconstrained problem with an extended-real-valued function $\left\Vert\mathbf{x}-\mathbf{a}\right\Vert^2 + I_\text{box}(\mathbf{x})$, where $I_\text{box}(\mathbf{x})$ is zero for $-\mathbf{1}\le\mathbf{x}\le \mathbf{1}$ and infinity otherwise.
Target $\mathbf{t}$ can be found by clipping the elements of $\mathbf{a}$ into the range $[-1, 1]$.

{\bf Constrained problem 2. Standard simplex:}
This is another constrained problem that finds the closest discrete probability distribution from a given vector, i.e.,
$\min_{\mathbf{1}^T\mathbf{x} = 1,~\mathbf{x}\ge\mathbf{0}}\left\Vert\mathbf{x}-\mathbf{a}\right\Vert^2$.
Likewise, this can be reformulated as an unconstrained problem with $\left\Vert\mathbf{x}-\mathbf{a}\right\Vert^2 + I_\text{simp}(\mathbf{x})$ where $I_\text{simp}(\mathbf{x})$ is analogously defined as in the previous problem.
The desired solution can be calculated based on~\cite{projection}.

Here, we again emphasize the fact that our method is not directly affected by any information about the functional relation or constraint. The only information being used is the training samples $(\mathbf{a}, \mathbf{t})$.

All the toy examples were evaluated under the following settings: Training and test sample inputs were randomly generated by zero-mean Gaussian distribution with standard deviation $2$ and $1$, respectively.
For training, 10,000 random samples were generated in every epoch. For test, 1,000 samples were generated and fixed for the entire procedure. The dimensions of $\mathbf{x}$, $\mathbf{a}$, and $\mathbf{t}$ were 10, and the size of mini-batch was 100.
For all problems, mean squared error (MSE) was used as the loss metric $L$.
The step size $\gamma$ was set to $10^3$ and the number of gradient steps $N$ to 1,000.
We used a simple two-layer network, i.e.,
\begin{center}
[vector input] - ($20\times 64$ FC) - (ReLU) - ($64\times 64$ FC) - (MS) - [scalar output].
\end{center}
Here, the input is the concatenation of $\mathbf{x}$ and $\mathbf{a}$, and FC and MS represent the fully-connected and the mean square operations, respectively.
This structure was applied to every toy example and the Adam~\cite{adam} optimizer was used in the outer process.

Figure~\ref{fig2} shows the experimental results for the toy examples.
Here, we can confirm that the losses of each problem decrease nicely along the training epochs as well as the gradient descent steps.
The shape of a learned objective function was drawn by perturbing the variable input near the target for a fixed sample input.
We randomly picked $10$ directions for perturbation and displayed the average and standard deviation of the objective function for each length of perturbation. Here, the figure shows that the learned objective functions have ideal shapes for optimization.

\subsection{Optical Flow}
In this subsection, we discuss the experimental results on optical flow estimation.
For the initialization of $u_0$ and $v_0$, we used one of the state-of-the-art optical flow methods, PWC-net~\cite{pwcnet}, implemented on PyTorch.
U-net~\cite{unet} with depth $3$ was used as the feature extraction network described in Section~\ref{3.3}, which gives the effect of using a $3$-level image pyramid.
The step size was set to $\gamma=1.0$, the regularization constant to $\lambda=180$, and the mini-batch size to $h=10$.

\begin{wrapfigure}{r}{0.45\linewidth}
  \vspace{-0.3cm}
  \includegraphics[width=1\linewidth]{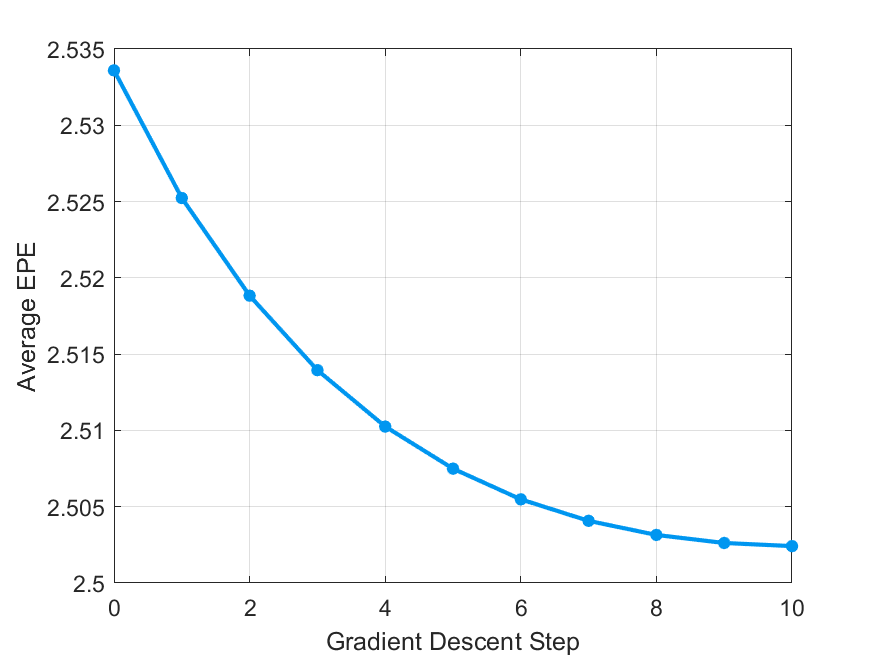}

  \caption{
  \textbf{Result for Optical Flow} Average EPE for test set decreases through the gradient steps.}
  \label{fig3}
  \vspace{-0.3cm}
\end{wrapfigure}

We used the MPI-Sintel \emph{clean} dataset~\cite{sintel} which is a 3D animation dataset for optical flow. 
Its training dataset consists of $23$ scenes, $1,064$ image frames, and $1041$ ground truth flows, which has been additionally separated to a training set ($908$) and a validation set ($133$) in an existing work~\cite{flownet}. In our experiment, this validation set was used as the test set because the original test set does not have any ground truth.

The average EPE of PWC-net for the test set is $2.534$ which is the initial error of $(u_0, v_0)$ in our algorithm.
We set the number of steps $N$ to $10$ which was the maximum possible number we could afford in the GPU memory. The 10 iteration steps improved the average EPE to $2.502$ as shown in Figure \ref{fig3}. The average EPE decreases smoothly as the iteration proceeds, which suggests that this improvement is not accidental and the proposed method learns a meaningful objective function. If future advances in deep learning hardware allow more memory space, than there is a chance that better performance can be achieved with more gradient descent steps.
\section{Conclusion}

In this paper, we proposed a novel framework called Neuro-Optimization which learns the objective function by a neural network and obtains the solution by optimizing it.
Unlike other methods where the network output becomes the solution, the input variable optimized by gradient descent becomes the solution in the proposed method, and the network parameters are updated to reduce the loss between the optimized result of the input variable and corresponding desired solution.
This allows us to learn a good objective function without heuristics or a number of hyper-parameters to be fine-tuned.
Experiments show that the proposed method can indeed learn valid objective functions, and we have demonstrated by the optical flow example that it can be successfully applied in real applications.
Our approach can be adopted in various problems, and there are many possible modifications to make. Especially, with the advance of computation power and memory space, we can expect better results by applying the proposed method in a more large-scale.

\subsubsection*{Acknowledgments}

This work was supported in part by the Next-Generation ICD Program through NRF funded by Ministry of S\&ICT [2017M3C4A7077582], and in part by the Basic Science Research Program through the National Research Foundation of Korea funded by the Ministry of Science and ICT under Grant NRF-2017R1C1B2012277.

\bibliographystyle{icml}
\bibliography{refs}

\end{document}